\documentclass[conference,10pt]{IEEEtran}
\usepackage[shortcuts,acronym]{glossaries}

\newacronym{ml}{ML}{Machine Learning}
\newacronym{dl}{DL}{Deep Learning}

\newacronym{ai}{AI}{Artificial Intelligence}
\newacronym{cam}{CAM}{class activation mapping}
\newacronym{grad-cam}{Grad-CAM}{gradient class activation mapping}
\newacronym{xai}{XAI}{Explainable Artificial Intelligence}
\newacronym{lar}{LAR}{Label-Aware Ranked}
\newacronym{shap}{SHAP}{SHapley Additive exPlanations}
\newacronym{lrp}{LRP}{Layer-wise Relevance Propagation}
\newacronym{gcn}{GCN}{Graph Convolutional Network}
\newacronym{fcn}{FCN}{Fully Connected Network}
\newacronym{dnns}{DNNs}{Deep Neural Networks}
\newacronym{bn}{BN}{Batch Normalization}
\newacronym{deeplift}{DeepLIFT}{Deep Learning Important FeaTures}
\newacronym{mp}{MP}{Max Pooling}
\newacronym{gap}{GAP}{Global Average Pooling}
\newacronym{dml}{DML}{Deep Metric Learning}
\newacronym{sm}{SM}{SHAP value map}
\newacronym{cv}{CV}{Computer Vision}
\newacronym{nlp}{NLP}{Natural Language Processing}

\newacronym{fmcw}{FMCW}{Frequency-Modulated Continuous-Wave}
\newacronym{adc}{ADC}{Analog to Digital Converter}
\newacronym{mti}{MTI}{Moving Target Indication}
\newacronym{fft}{FFT}{Fast Fourier Transform}
\newacronym{rdi}{RDI}{Range Doppler Image}
\newacronym{rai}{RAI}{Range Angle Image}
\newacronym{cnn}{CNN}{Convolutional Neural Network}
\newacronym{dgcnn}{DGCNN}{Dynamic Graph CNN}
\newacronym{doa}{DOA}{Direction Of Arrival}
\newacronym{mse}{MSE}{Mean Squared Error}
\newacronym{lstm}{LSTM}{Long short-term memory}
\newacronym{rcnn}{R-CNN}{Regions with \ac{cnn} features}
\newacronym{mlp}{MLP}{Multi-Layer Perceptron}
\newacronym{if}{IF}{Intermediate Frequency}
\newacronym{gans}{GANs}{generative adversarial networks}
\newacronym{ewc}{EWC}{elastic weight consolidation}
\newacronym{si}{SI}{synaptic intelligence}

\usepackage{lipsum}
\usepackage{todonotes}
\usepackage[english]{babel}               
\usepackage{amsmath}

\usepackage{amsfonts}
\usepackage{graphicx}
\usepackage{subcaption}
\usepackage[font=small,labelfont=bf]{caption} 
\usepackage{booktabs}
\usepackage{multirow}
\usepackage{array}
\usepackage{algorithm}
\usepackage{algorithmic}
\usepackage{hyperref}
\usepackage{tikz}
\usepackage{pgfplots}
\usepgfplotslibrary{groupplots}
\usepackage{float}
\usepackage{epstopdf}
\usepackage{enumitem}
\usepackage{setspace}
\usepackage{nicefrac}
\usepackage{url}

\usepackage{bm}
\usepackage{mathtools, nccmath}
\usepackage{svg}
\usepackage{pdfpages}

\pgfplotsset{compat=1.17}
\begin{document}
\IEEEpubid{\makebox[\columnwidth]{0000--0000/00\$00.00 $\copyright$2015 IEEE \hfill} \hspace{\columnsep}\makebox[\columnwidth]{ }}
\title{Utilizing Explainable AI for improving the Performance of Neural Networks}
\author{\IEEEauthorblockN{{Huawei Sun$^{1,3}$, Lorenzo Servadei$^{1,3}$, Hao Feng$^{1,3}$, Michael Stephan$^{1,2}$, Robert Wille$^{3}$,  Avik Santra$^{1}$}}
\IEEEauthorblockA{\textit{$^{1}$Infineon Technologies AG}, Neubiberg, Germany\\
$^{2}$Friedrich-Alexander-University Erlangen-Nuremberg, Erlangen, Germany\\
$^{3}$Technical University of Munich, Munich, Germany\\
E-mail: \{huawei.sun, lorenzo.servadei, avik.santra\}@infineon.com\\
\{hao.feng, robert.wille\}@tum.de\\
\{michael.stephan\}@fau.de}}


\maketitle
\begin{abstract}
 Nowadays, deep neural networks are widely used in a variety of fields that have a direct impact on society. Although those models typically show outstanding performance, they have been used for a long time as black boxes. To address this, \ac{xai} has been developing as a field that aims to improve the transparency of the model and increase their trustworthiness. We propose a retraining pipeline that consistently improves the model predictions starting from \ac{xai} and utilizing state-of-the-art techniques. To do that, we use the \ac{xai} results, namely \ac{shap} values, to give specific training weights to the data samples. This leads to an improved training of the model and, consequently, better performance. In order to benchmark our method, we evaluate it on both real-life and public datasets. First, we perform the method on a radar-based people counting scenario. Afterward, we test it on the CIFAR-10, a public Computer Vision dataset. Experiments using the SHAP-based retraining approach achieve a 4\% more accuracy w.r.t. the standard equal weight retraining for people counting tasks. Moreover, on the CIFAR-10, our SHAP-based weighting strategy ends up with a 3\% accuracy rate than the training procedure with equal weighted samples.  
\end{abstract}
\section{Introduction}
Various application areas have been positively affected by the recent advances of \ac{ai} and \ac{ml}. Among them, fields such as autonomous driving \cite{you2019advanced, grigorescu2020survey}, health tech \cite{torres2018patient,chen2020deep} and robotics \cite{robotics1, robotics2} heavily rely on the processing of \ac{ml} algorithms onto a set of different sensors. These approaches are typically based on computationally intensive \ac{dl} strategies, which involve training millions, or even billions of parameters to perform a specific task. Although the out-coming results show high performance, a major problem occurs: As a neural network gets deeper and deeper, it is also becoming more complex and thus challenging to be interpreted. To this end, a neural network is often considered a black box: Even if the model correctly predicts the given specific input,  it is difficult to explain what causes the correct prediction. This property, in turn, reduces the trustworthiness of the outcome.

In order to improve a \ac{dl} system, it is necessary to understand its weaknesses and shortcomings \cite{samek2017explainable}. Approaching this, \ac{xai} focuses on improving the transparency of \ac{ml} technologies and increasing their trust. When a model's predictions are incorrect, explanatory algorithms can aid in tracing the underlying reasons and phenomenon. \ac{xai} has been researched for several years, and lots of work has been done in  fields such as \ac{cv} \cite{cam, gradcam, lime} and \ac{nlp} \cite{nlp1, nlp2}. These algorithms mainly generate attention maps, which help to highlight the critical area/words in classifying images or during language translation.
Nevertheless, nowadays \ac{dl} is widely applied in less conventional application fields: For example, radar-based solutions for tasks such as counting people \cite{ppl_michael}, identifying gestures \cite{gesture}, and tracking \cite{tracking}, as shown in this contribution \cite{radar_book}.
Although the advancements mentioned above successfully solve radar-based problems, explaining \ac{dl} models for radar signals is still a challenging topic. Additionally, most \ac{xai} algorithms analyze the predictions from a well-trained model, thus focusing only on the explanatory part. To this end, a few research contributions move forward by utilizing the results from \ac{xai} for secondary tasks. \ac{lrp}, for example, is used for adaptive learning rate during training in \cite{xai_lr}, and in \cite{xai_prun} the authors prune \ac{dnns} and quantize the weights mainly by \ac{deeplift}. 
However, to the best of our knowledge, \ac{xai} has not yet been used to process the dataset and improve the network performance. In this paper, we first adapt our method to a real-life use case: Radar-based people counting. Afterward, we show promising results on the CIFAR-10 dataset \cite{cifar} to further underlying the approach's generality.

Radar-based people counting is a significant application with high privacy preservation and weather condition independence compared to camera-based people counting. However, the algorithms often underperform the state-of-the-art computer vision methods, and radar data often has the limitation of low-resolution and room dependency \cite{ppl1,ppl2}. To this end, many solutions have been implemented which use \ac{dl} for this task in different scenarios \cite{ppl_michael,lar}. Although performant, those solutions do not consider how the network obtains the prediction and which features are essential to explain the outcome of the task.

This paper introduces a retraining pipeline, which adopts the \ac{shap} values for improving the model performance. The proposed approach shows convincing outcomes in both public and real-life datasets. In the radar-based people counting scenario, on the one hand, we explain the neural network's prediction of radar signals. This is the first time in the literature that \ac{xai} algorithms are applied to radar-based input neural networks. On the other hand, we propose a retraining pipeline that adds sample weights generated from the \ac{shap} \cite{shap} results. 
This further improves the performance of the people counting network. 
To show this, we execute several experiments highlighting our method's benefit. We first apply our weighting strategy to our radar-based people counting task.
Compared with the equal weighting method, our SHAP-based method ends up with an increase of $4\%$. When we apply the method to CIFAR-10, our approach outperforms default equal weighting retraining with a $3\%$ more accuracy.



\section{Background and Motivation}
\label{background}
This section first describes the status of explainable AI and its application fields. Afterward, we introduce the fundamentals of mmWave \ac{fmcw} radar.

\subsection{Explainable Artificial Intelligence}
Several \ac{cam} based \ac{xai} algorithms such as \ac{cam} \cite{cam} and \ac{grad-cam} \cite{gradcam} are used, in the literature, to explain well-trained CNN architecture in image classification tasks. They typically work by generating saliency maps through a linear combination of activation maps that highlight the model's attention area. Although those methods are widely-spread, there is still a lack of research on using \ac{xai} algorithms to explain models with radar signals as input. In fact, on the one hand, features in radar signals are more difficult to understand for humans than in images. On the other hand, unlike image RGB channels, radar signals are typically represented by distinct information, such as Macro- and Micro-Doppler maps which cannot be stacked and treated as one single image. This contrasts with the usual practice in CV, where methods such as \ac{cam}-based approaches only generate a single saliency map for features of each input sample. Therefore, a major question is: how to utilize an efficient \ac{xai} method adaptable to the heterogeneous information contained in the input. Recently,  \ac{shap}\cite{shap} has been used in domains such as text classification \cite{shap_text} or analysis of time-series data \cite{shap_time}, as an additive feature attribution approach, which can explain multiple information in the input at the same time. This exactly adapts to the problem at hand.

\begin{description}[leftmargin=0pt]
    \item[Additive feature attribution method]
    Additive feature attribution method follows:
    \begin{equation}
    \label{eq:additive}
        g\left(z'\right) = \phi_{0} + \sum_{i=1}^{M} \phi_{i} z_{i}'
    \end{equation}
    \noindent where $z' \in \{0, 1\}$  is a binary vector whose entries are the existence of the corresponding input feature and $M$ is the number of simplified input features. $\phi_{i}$ indicates the importance of the $i^{th}$ feature and $\phi_{0}$ is the baseline explanation.
    In order to explain a complex model $f$, such as a neural network, we can use a more straightforward explanation model $g$ instead. For a given input $x$, $f(x)$ denotes the prediction output. A simplified input $x'$ can be restored to the input $x$ through a mapping function $x=h_{x}(x')$. When the binary vector $z'\approx x'$, additive feature attribution methods ensure $g(z')\approx f^{c}(h_{x}(x'))$ where $c$ is the target class of given input $x$.
    
    \item[\ac{shap} values] \ac{shap} values are the feature attributions of the explanation model, which obeys Eq. \ref{eq:additive} and are formulated as follows:
    \begin{equation}
    \label{eq:shap}
    \phi_{i} = \sum_{z' \subset x'}\frac{(M-|z'|)!(z'-1)!}{M!}[f^{c}(h_{x}(z'))-f^{c}(h_{x}(z'\backslash i))]
    \end{equation}
    \noindent where $|z'|$ denotes the number of non-zero entries in $z'$ and $z' \subset x'$ represents all $z'$ vectors where the non-zero entries are a subset of the non-zero entries in $x'$. Additionally, $z'\backslash i$ denotes setting $z_{i}$ to zero. In this way, taking image-based input as an example, for every single model input, it can generate one \ac{shap} map in the same shape as the input sample, where the pixel value denotes attribution of the corresponding pixel from the input.

\end{description}


\subsection{Introduction of mmWave \acrshort{fmcw} Radar Sensor}
\label{radar signal processing}
\ac{fmcw} radar sensors typically operate by transmitting a sequence of modulated frequency chirp signals with a short ramp time and delays between them. A chirp sequence with a fixed number of chirps is usually defined as one frame and repeated with the frame repetition time \cite{ppl_michael}. 
The chirp signals reflected by the
targets are received, mixed with the transmitted signal, and filtered, to generate the \ac{if} signal, which is then digitized by the \ac{adc} for task-dependent pre-processing. For example, range and Doppler information can be acquired by taking the \ac{fft} along the respective axes of the data in one frame. Using multiple receiving or transmitting antennas also allows the estimation of the \ac{doa} from the \ac{doa}-dependent time delay of the received signal across the antennas. This makes \ac{fmcw} radar the most commonly used since it avoids complicated pre-processing and saves energy at the same time. Therefore, it is an ideal sensor for many simple tasks, such as people counting and activity classification, without losing users' data privacy.


\section{Approach}
\label{approach}
In this section, we first describe the radar data preprocessing method. Then, we illustrate the radar data augmentation methods. We also introduce a stabilized architecture for learning robust embedding vectors and label predictions. Ultimately, we focus on our retraining procedure with different weighting methods calculated from \ac{shap} values and probability vector predictions. 
\subsection{Radar Data Preprocessing}

\begin{figure}[]
    \centering
    \includegraphics[width=0.85\linewidth]{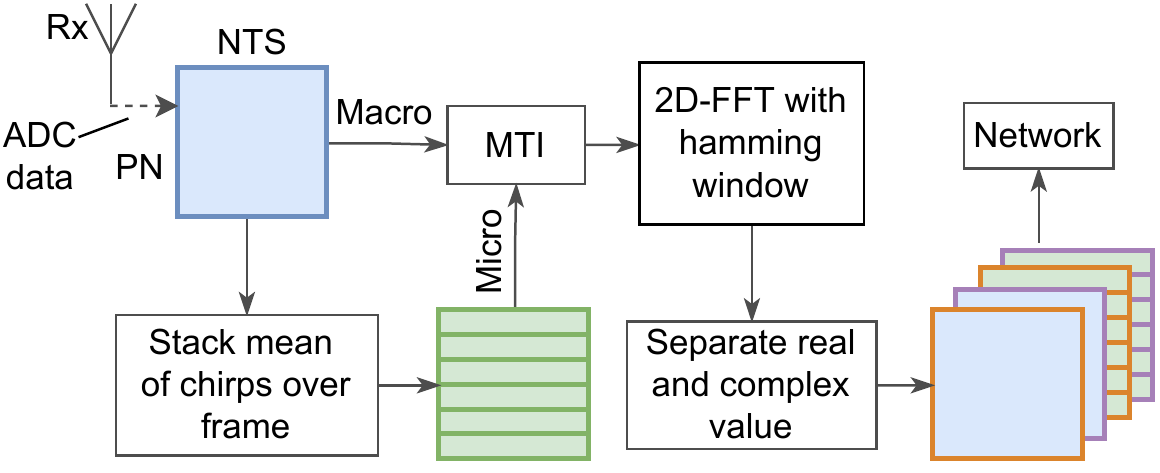}
    \caption{Radar Pre-processing.}
    \label{fig:radar_prep}
    \vspace{-4mm}
\end{figure}

\label{preprocessing}
The radar signal preprocessing pipeline is illustrated in Fig.\ref{fig:radar_prep}.
The acquired \ac{if} signals form a raw data frame in the shape of $PN \times NTS$ in each Rx antenna channel, where $PN$ and $NTS$ are the number of chirps per frame and number of samples per chirp, respectively. It is first passed through a \ac{mti} filter to remove static objects and Tx-Rx leakage effects by subtracting the mean value along the chirp axis (slow time). Then, range-\ac{fft} and doppler-\ac{fft} are applied along fast time (sample axis) and slow time respectively to generate 2D \acp{rdi}. Additionally, before the respective \acp{fft}, we multiply the data with the hamming window of the corresponding sizes to suppress side lobes that cause leakage into the adjacent \ac{fft} bins. Meanwhile, to obtain Micro Doppler information in higher Doppler resolution, the observation time is extended by stacking the mean of chirps in each frame across $PN$ consecutive frames to compute an additional \ac{rdi} with the same shape as the previous Macro-\ac{rdi}. Afterward, we divide each acquired \ac{rdi} into real and imaginary values and stack them together. We finally get a 3D matrix in the shape of $PN \times NTS/2 \times 4$ for each antenna's data in each frame, including Macro- and Micro- real and imaginary values.

\subsection{Radar Data Augmentation}
\begin{figure}
    \begin{subfigure}[b]{.48\columnwidth}
        \includegraphics[width=\linewidth]{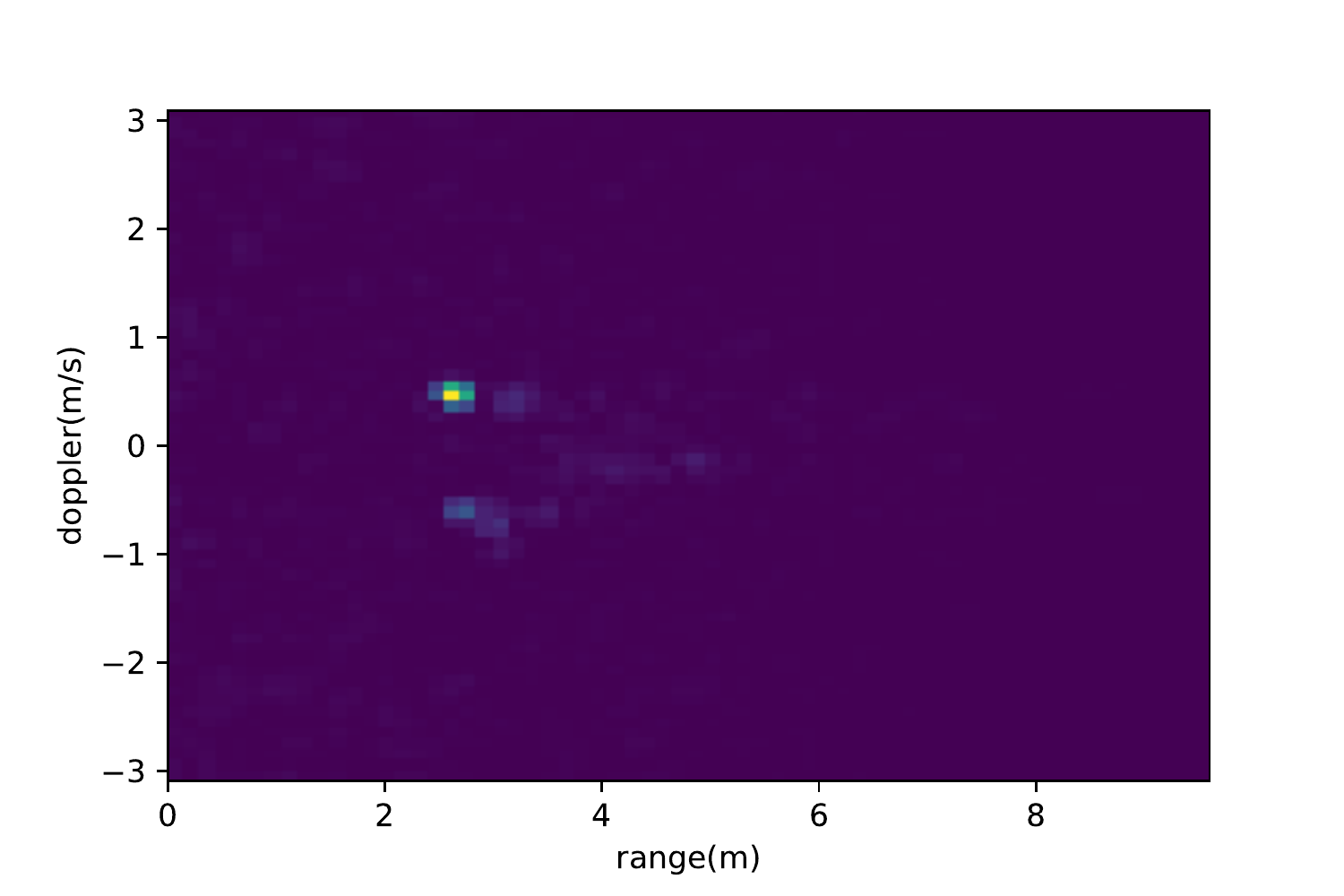}
        \caption{Original \ac{rdi}.}
        \label{fig:original_rdi}
    \end{subfigure}
    \hfill
    \begin{subfigure}[b]{0.48\columnwidth}
        \includegraphics[width=\linewidth]{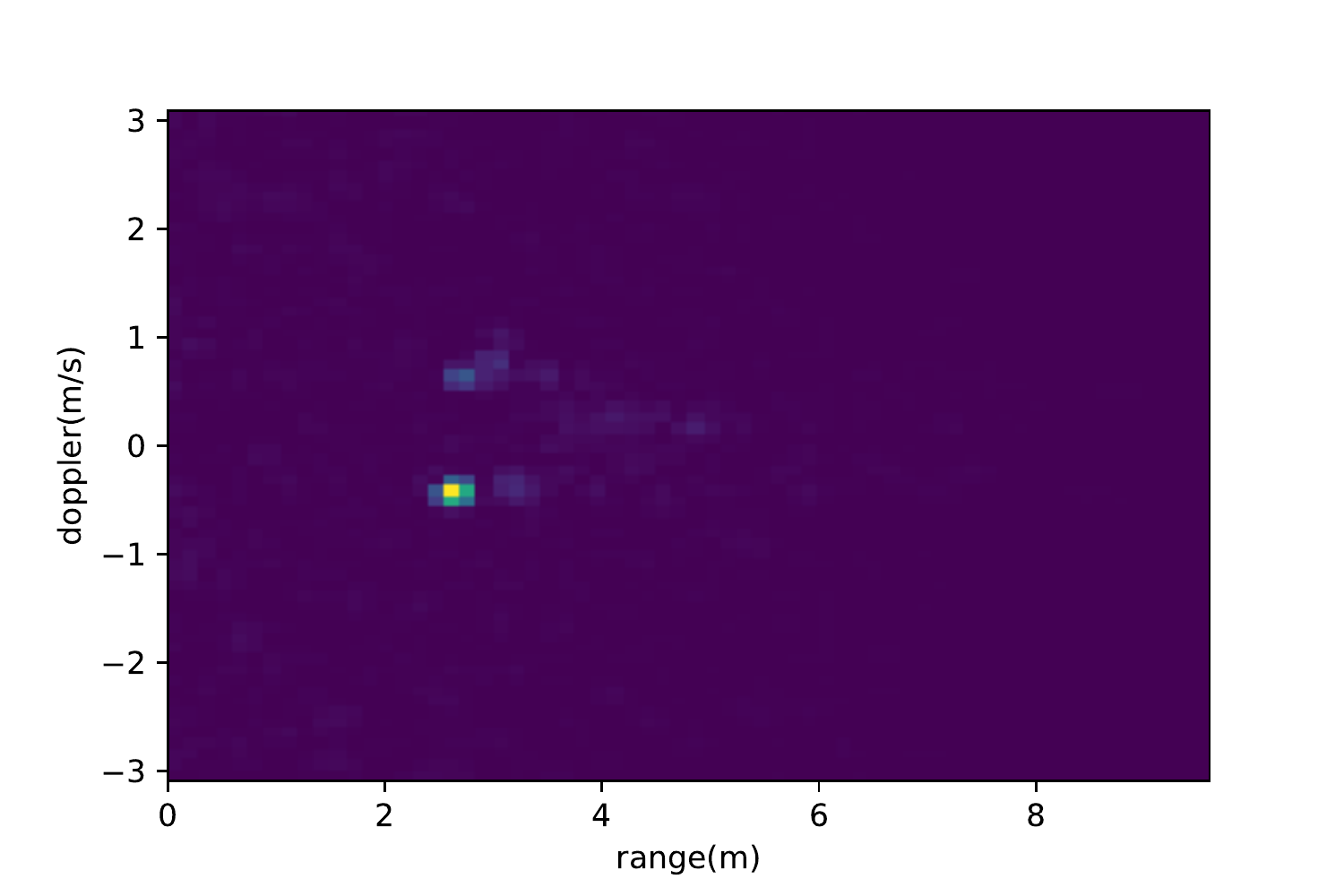}
        \caption{Flipped \ac{rdi}.}
        \label{fig:flip}
    \end{subfigure}
    \hfill
    \begin{subfigure}[b]{0.48\columnwidth}
        \includegraphics[width=\linewidth]{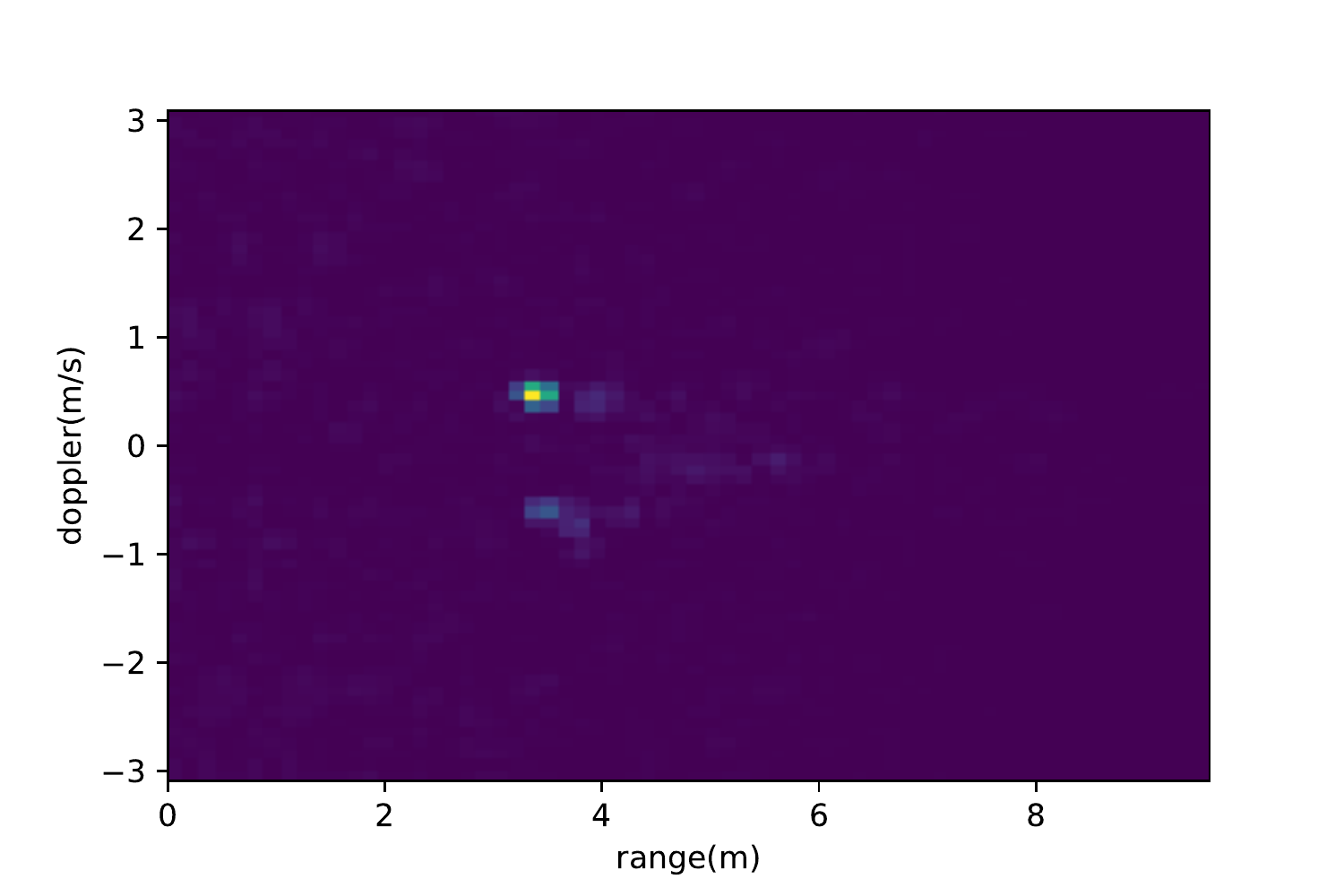}
        \caption{Shifted \ac{rdi}.}
        \label{fig:shift_rdi}
    \end{subfigure}
    \hfill
    \begin{subfigure}[b]{0.48\columnwidth}
        \includegraphics[width=\linewidth]{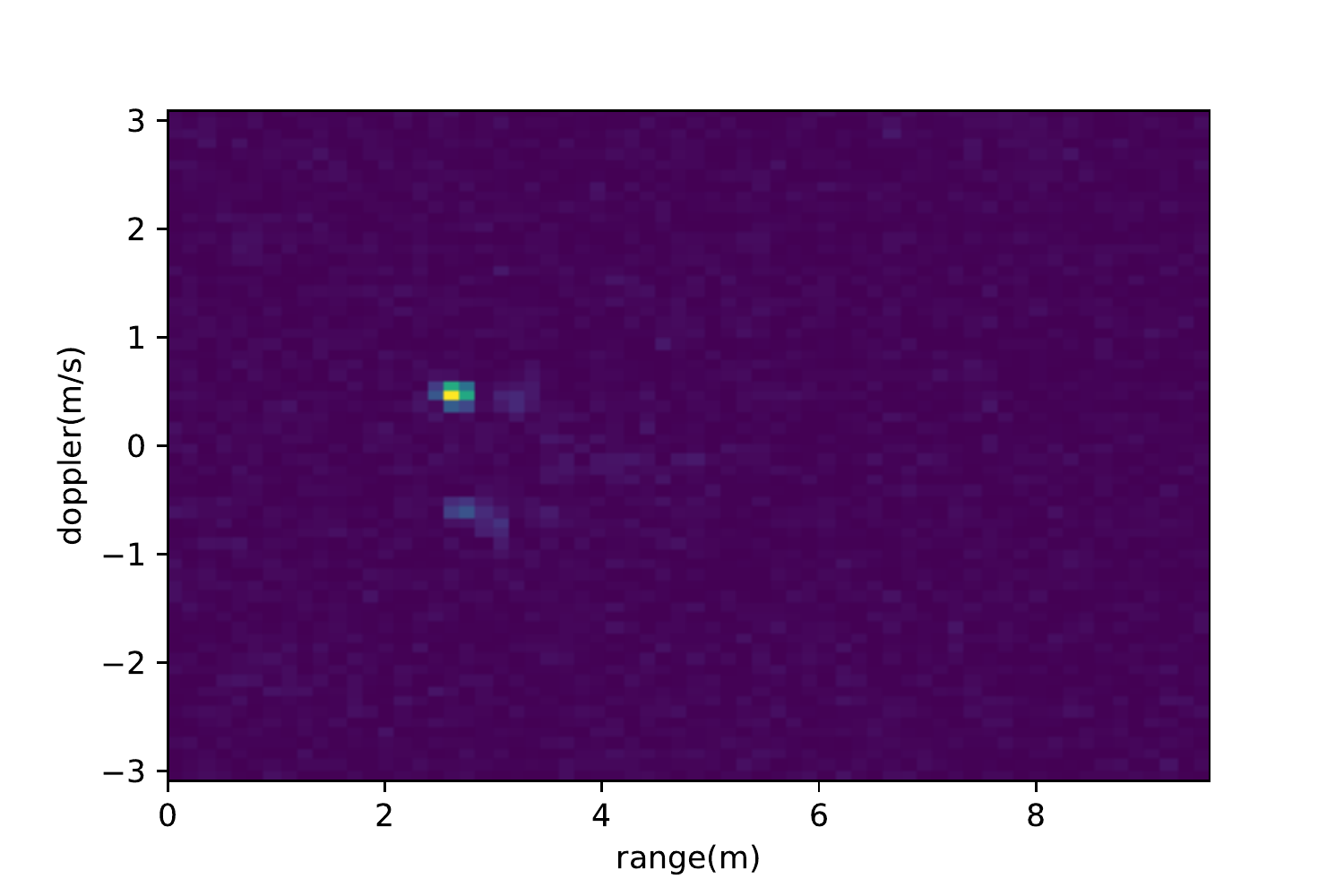}
        \caption{Add Gaussian noise ($\sigma=0.01$).}
        \label{fig:add_noise}
    \end{subfigure}
\caption{Data augmentation methods.}
\label{fig:augmentation}
\vspace{-4mm}
\end{figure}

The robustness of the model benefits from training with augmented data. However, we must ensure that the augmented radar data maintains its kinematic properties. This paper proposes two data augmentation methods, as shown in Fig. \ref{fig:augmentation}. 

\subsubsection{Frequency Shift}
As mentioned in Sec. \ref{preprocessing}, after applying FFT twice to the raw data, we get \ac{rdi} as input data. Precisely, the horizontal axis measures the range between the moving target and the radar, and the vertical axis is the target velocity. Here, the \ac{rdi} denotes $x$, and the pixel on the \ac{rdi} is indicated as $x(r, d)$. Considering the kinematic property of \ac{rdi}, we apply two kinds of frequency shift augmentation methods to the original \ac{rdi}: 

\begin{description}[leftmargin=0pt]
    \item[Flip along Doppler] As shown in Fig. \ref{fig:augmentation}, moving targets can have positive or negative velocity. Assuming the target is moving towards the radar in the people counting scenario, the velocity is negative. Flip the \ac{rdi} along the doppler axis also keeps its kinematic property when we assume that the target is moving away from the radar after flipping.
    \begin{equation}
        x_{aug}(r,d) = x(r, \frac{NTS}{2}-d)
    \end{equation}

    \item[Shift along Range] The horizontal axis measures the distance between the radar and the moving object. Since, during recording, several people are moving in the room, we need to consider the minimum distance from the one closest to the radar to specify the maximum moving distance parameter as $R$. We randomly choose a shifting range $R_{s}$ for each frame from $0$ to $R$. The range bin within $R_{s}$ of the \ac{rdi} remains unchanged.
    \begin{equation}
        x_{aug}(r,d) = \left\{ \begin{array}{rcl}
            x(r,d) & \mbox{for} & r<R_{s} \\
            x(r-R_{s}, d) & \mbox{for} & R_{s}\leq r \leq \frac{NTS}{2}
        \end{array} \right.
    \end{equation}
    
\end{description}


\subsubsection{Adding Gaussian Noise}
Adding noise can help prevent the network from overfitting and protect the network from adversarial attacks \cite{gaussian}. Since radar data are more sensitive to noise, if the variance of the Gaussian noise is too high, the intensity value in the \ac{rdi} changes a lot. An example of the \ac{rdi} with noise shows in Fig. \ref{fig:add_noise}. On the other hand, directly training the network by adding noisy inputs takes longer to converge and leads to underfitting if the hyperparameters are not chosen appropriately. That is why we introduce a \textbf{Stability Training Architecture} which is explained in detail in the next section.

\subsection{Stabilized Architecture}
\begin{figure}
     \centering
     \includegraphics[width = 0.85\linewidth]{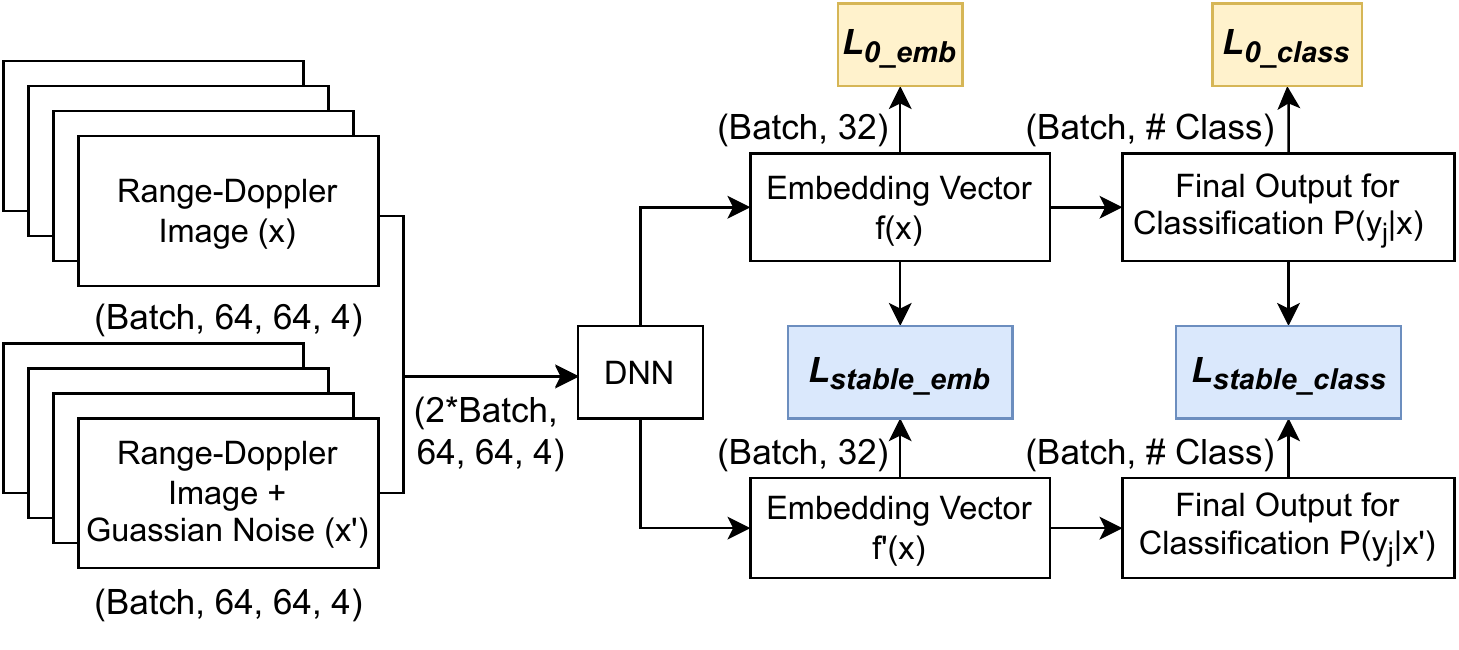}
     \caption{Stability Training Architecture.}
     \label{Fig: stability}
     \vspace{-4mm}
 \end{figure}

As \ac{dl} is used in more and more areas, there is a growing demand for model robustness. Recently, researchers have found that a small perturbation added to the input in image classification tasks can lead to a highly wrong prediction from the well-trained model, such as studies of adversarial attack in \cite{adversarial,adversarial2}. 
Even state-of-the-art neural networks have trouble avoiding this, so model-robustness becomes a hot and challenging topic. 
However, in using \ac{dl} to analyze radar signals, little research has been done to improve the robustness \cite{radar_adversarial,radar_adversarial2}. 
Similar to the contribution of \cite{stability} for Computer Vision, we introduce training with the stability loss, which we name \emph{stability training}, for the radar application. The architecture is shown in Fig. \ref{Fig: stability}. 

While doing the stability training, the original \ac{rdi}s are used as input $x$ with batch size $N_{B}$. Afterward, we create a perturbed version of $x$ as $x'$ by adding pixel-wise uncorrelated Gaussian noise $\epsilon$. If $x_{i}$ is the $i^{th}$ pixel of $x$, then $x_{i}'$ is described in the following:
\begin{equation}
    x_{i}' = x_{i} + \epsilon_{i}, \: \epsilon_{i} \sim \mathcal{N}(0, \sigma_{i}),  \: \sigma_{i}>0
\end{equation}
\noindent where $\sigma_{i}^{2}$ is the variance of the Gaussian noise, which we add manually to the $k^{th}$ pixel of $x$. 


After creating the noisy version of the input $x$ and adding it to the input data, the batch size becomes $2N_{B}$. Passing the concatenated input through the DNN block, we get an embedding vector with size $(2N_{B}, 32)$ and a final output vector with size $(2N_{B}, L)$, where $L$ is the number of different labels. 

\subsubsection{Stability for embedding vectors}
We apply \ac{dml} on embedding vectors $f(x)$, aiming to cluster samples from the same class, and dissimilar pairs from different classes should fall far apart in the embedding space. Here, we use \ac{lar} loss \cite{lar} instead of the hinge loss between triplets since the \ac{lar} loss is a more precise \ac{dml} loss for the people counting task, which ranks the embedding space by its label number, as shown in the following equation:
    
\begin{equation}
\label{eq:lar}
\resizebox{0.48\textwidth}{!}{$L_{0\_emb}=\frac{1}{N}\sum_{i=1}^{N}\log(1+\sum_{j\neq i}\exp (\log(\Delta l)f_{i}(a)f_{j}(n)^{T}-f_{i}(a)^{T}f_{j}(p)))$}
\end{equation}
where
\begin{equation}
\label{eqn:multipl}
\Delta_l=min\left(|l_a - l_n|, \left|L-|l_a-l_n|\right| \right)
\end{equation}
In Eq. \ref{eq:lar}, we select triplet pairs \cite{triplet} ${f(a), f(p), f(n)}$ from embedding vectors where $f(a)$ and $f(p)$ have the same label $l_{a}$ and $f(n)$ has a different label $l_{n}$.
    
In addition, we also require the noisy embedding vector $f(x')$ to be as close as possible to the original embedding vector $f(x)$ to ignore the impact by adding noise. This helps to decrease the distance between $f(x)$ and $f(x')$ in the embedding space by minimizing the $L2$-distance between these two vectors:
\begin{equation}
    \label{eq:l2}
    L_{stable\_emb} = \Vert f(x)-f(x') \Vert_{2}
\end{equation}

\subsubsection{Stability for classification}
In the classification stage, we also need to minimize two losses. Here, $P(\hat{\textbf{y}}|x)$ denotes the output probability vector for a given input $x$, and $\textbf{y}$ represents the ground truth binary label vector. We train the network by minimizing standard cross-entropy loss for the original input $x$:
\begin{equation}
    L_{0\_class} = -\sum_{j}y_{j} \log P(\hat{y_{j}}|x).
\end{equation}
To use the stability training for classification, we need $P(y_{j}|x')$ to be similar to $P(y_{j}|x)$. In this case, we use KL-divergence to realize the requirement:
\begin{equation}
    L_{stable\_class} = -\sum_{j}P(\hat{y_{j}}|x) \log P(\hat{y_{j}}|x')
\end{equation}

Finally, we sum up these four above losses with equal weight as stability loss.
\begin{equation}
    Loss=L_{0\_emb}+L_{stable\_emb}+L_{0\_class}+L_{stable\_class}
\end{equation}
and is used for \textit{stability training}.



\subsection{Retraining Methodology}
\label{weight}
In this paper, we introduce our retraining pipeline to improve model performance.
The retraining procedure is shown in Alg. \ref{algo}. In the baseline training procedure, we are given a set of data $D^{m}$ for training the baseline network, a validation set $D^{v}$ for validating during training, and an untouched test set $D^{t}$ for evaluating the network performance when the training session is completed. After the baseline network training session, we create several evaluation sets $D^{v_{i}} \ldots D^{v_{s}}$, in order to explore the impact of new data on the network performance.
In our framework, to generate new data, we select the wrong predicted samples out of the evaluation sets, and create incremental datasets $D^{I_{1}} \ldots D^{I_{s}}$ for the next retraining session. As an example, in the $1^{st}$ retraining session, the training dataset becomes $D^{m} \cup D^{I_{1}}$. We continue to retrain the network $s$ times by using datasets $D^{m} \cup D^{I_{1}} \cup \ldots D^{I_{s}}$ with the unchanged weighting parameter $w$.

\begin{algorithm}
\caption{Retraining Procedure}
\label{algo}
\begin{algorithmic}[1]
 \REQUIRE $D^{m}, D^{v}, D^{t}$
 \\ \textit{Baseline Training} :
  \STATE Initialize model $\theta^0$
  \STATE $\theta^{m}=train(D^{m},\theta^{0},D^{v})$
  \STATE $Acc^{m}=eval(\theta^{m},D^{t})$
 \\ \textit{Generate \ac{shap} Value}
  \FOR {$D^{v_i}$ in $[D^{v_1},\dots,D^{v_s}]$}
  \STATE $D^{I_i}=\{(d,y) \in D^{v_i} | pred(\theta^{m},d)\neq y\}$
  \STATE $\ac{shap}^{I_i}=shap(\theta^{m},D^{I_i})$
  \STATE $w^{I_i}=Weight\_Cal(\ac{shap}^{I_i})$
  \ENDFOR
 \\ \textit{Incremental Training}
  \STATE Initialize $D=D^{m}, \theta^{0}=\theta^{m},w=1$
  \FOR {$D^{I_i}$ in $[D^{I_1},\dots,D^{I_s}]$}
  \STATE \text{Add new data:} $D=D\cup D^{I_i}$, $w=w\cup w^{I_i}$
  \STATE $\theta^{i}=train(\theta^{i-1},D,w,D^{v})$
  \STATE $Acc^{i}=eval(\theta^{i},D^{t})$
  \ENDFOR
\end{algorithmic} 
\end{algorithm}

As an example, let $(\textbf{x},\textbf{y})$ be an input-target pair with batch size $N_{B}$, and  $x_{i}\in \mathbb{R}^{W\times H\times C}$ denotes the $i^{th}$ sample of width $W$, height $H$ and channel $C$. Let $\Phi(x,\theta)$ be our neural network with parameters $\theta$, and $L(\hat{y}, y)$ is the loss function to minimize during the training step. Here, $\hat{y}$ is the output of the model and $y$ is the ground truth label. Without any weighting parameters, we aim to minimize the expected loss: $\frac{1}{N_{B}}\sum_{i=1}^{N_{B}}L(\hat{y_{i}},y_{i})$, where each input sample has equal weight.
The intuition of our retrain framework is the following: as the correct prediction on some data sample is more difficult to learn than for others (i.e., leads to incorrect estimations), weighting each input sample helps to focus on the most difficult predictions. As a consequence, we incorporate this principle in the following weighted loss: 
\begin{equation}
    \theta^{\ast}(w) = \mathop{\arg\min}_{\theta} \sum_{i=1}^{N_{B}}w_{i}L(\hat{y_{i}},y_{i}).
\end{equation}
In the equation, a higher weighting parameter $w_{i}$ gives higher importance to a specific input sample $x_{i}$ in backpropagation during the training session. 
In order to generate those weights, we introduce different methods for obtaining those parameters.
\subsubsection{Calculate from probability vector of prediction}
The most straightforward idea of giving weight to the input sample is that the wrong-predicted data would be assigned with a higher weight. $P(y|x_{i})$ denotes the prediction's probability vector for data classified incorrectly. Further, $p=\arg\max P(y|x_{i})$ and $l$ are the wrong prediction label and ground-truth label, respectively. Hence, during the retraining session, we set the weighting parameter $w_{i}$ as: 
\begin{equation}
    w_{i} = \left\{ \begin{array}{rcl}
            1 & \mbox{if} & p=l \\
            1+P(y_{p}|x_{i})-P(y_{l}|x_{i}) & \mbox{if} & p\neq l
        \end{array} \right.
\end{equation}

Next, we show how \ac{shap} values can help determine proper sample weighting parameters.
 
\subsubsection{Calculate from \ac{shap} value}
\begin{figure*}
    	\begin{center}
    	    \vspace{-0mm}
    	\includegraphics[width=0.9\textwidth]{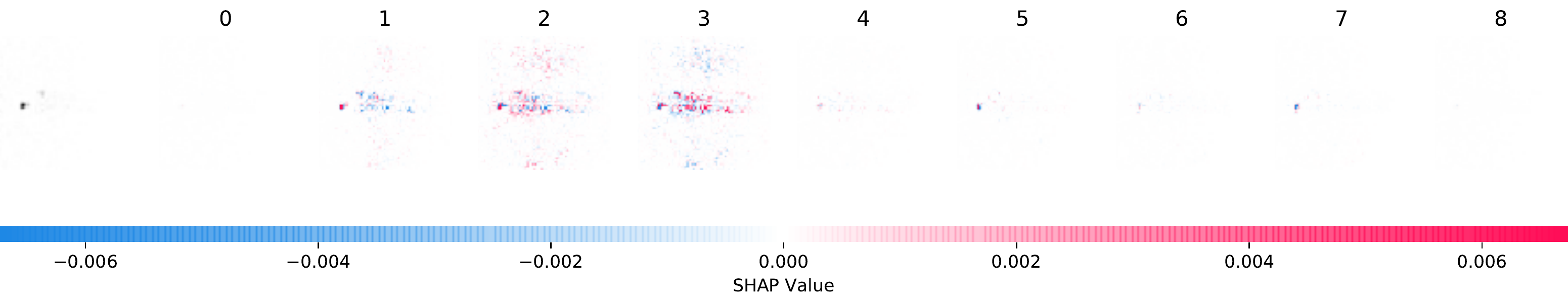}
    	\end{center}
    	\caption{\textbf{\ac{shap} value example of radar image.} This is the Macro-real map $\textbf{M}$ of one input sample $x_{i}$ which has the ground truth label 2 ($y_{i}=2$) and incorrectly classified as label 3 ($\hat{y_{i}}=3$). Features on map $M_{y_{i}}$ pushing the prediction as label $y_{i}$ higher are shown in red, those pushing the prediction lower are in blue.}
    	\vspace{-4mm}
    	\label{fig:shap}
\end{figure*}

\ac{shap} values show how pixels contribute to the specific prediction class of a well-trained model. An example of \ac{sm} is shown in Fig. \ref{fig:shap}. Take the $n^{th}$ \ac{shap} value map $M_{n}$ as an example, $M_{n} \in \mathbb{R}^{W\times H}$ has the same shape as the input, and the pixels of \ac{sm} correspond to the pixels of the input image, respectively. The red part has a positive contribution for predicting as label $n$, and the negative value pixels contribute adversely to this prediction class. Each input channel generates $L$ \ac{sm}s. For a $C$ channels input image, it generates $C\times L$ \ac{sm}s for the explanation. 

During the baseline training stage, the training set has already overfitted to a 99\% accuracy rate, and incremental datasets contain only incorrect predicted data. So that in the retraining stage, we focus more on the incremental datasets and set the sample weighting parameter to 1 if the data comes from $D^{m}$. Assume now we are at the $1^{st}$ retraining stage and $x_{i}\subset D^{I_{1}}$. More specifically, $x_{ik}$ is the $k^{th}$ input channel where $k\in [0, C]$. Here, we introduce two methods for calculating the sample weight by \ac{sm}s which focus on a different perspective of incorrect predictions. $M_{ik}$ denotes the \ac{sm} for predicting $x_{ik}$ as ground truth label $y_{i}$ and $\hat{M_{ik}}$ represents the \ac{sm} for predicting $x_{ik}$ as incorrect prediction label $\hat{y_{i}}$. Furthermore, $M_{ik}^{p_{w},p_{h}}$ is the \ac{shap} value at the position $(p_{w},p_{h})$ on the specific \ac{sm}  $M_{ik}$ where $0\leq p_{w} \leq W-1$, $0\leq p_{h} \leq H-1$.

\begin{description}[leftmargin=0pt]
    \item[Masked difference] In this method, we concentrate on the ground truth label. Since the sample $x_{i}$ leads to the wrong prediction $\hat{y_{i}}$, a higher positive value in $\hat{M_{ik}}$ stands for a higher contribution of the specific pixel for predicting incorrectly. Although a negative value in $\hat{M_{ik}}$ means that this pixel is against predicting as label $\hat{y_{i}}$, it does not imply that it positively contributes to predicting the correct label. In this case, if we want the algorithm to concentrate more on pulling the input in the direction of a correct prediction, we mask those pixels. Here, $\mathcal{A}_{ik}$ denotes the subset which includes the position pairs $(p_{w},p_{h})$ that satisfy $\hat{M_{ik}^{p_{w},p_{h}}}>0$. We define $\mathcal{A}_{ik}$ as:
    \begin{equation}
        \mathcal{A}_{ik}=\{(p_{w},p_{h})|\hat{M_{ik}^{p_{w},p_{h}}}>0\}
    \end{equation}
    Moreover, we set the additional weighting parameter of input sample $x_{i}$ as:
    \begin{equation}
        \Delta w_{i}=\sum_{k} \frac{1}{ \mid \mathcal{A}_{ik}\mid}(\sum_{(p_{w},p_{h})\in \mathcal{A}_{ik}}(\hat{M_{ik}^{p_{w},p_{h}}}-M_{ik}^{p_{w},p_{h}}))
    \end{equation}

    
    \item[Localize difference] On the contrary, in this method, we focus on the prediction map. The goal here is to pull the prediction of this data sample out of the previous prediction $\hat{y_{i}}$. 
    After calculating the pixel-wise difference between $\hat{M_{ik}}$ and $M_{ik}$, we sum those up to create a distance map. The higher is the distance between the two maps, the more difficult it is for the network to classify the input data sample. Thus, a higher additional weighting parameter needs to be assigned. We set $\Delta w_{i}$ in the following:
    \begin{equation}
        \Delta w_{i} = \sum_{k}(
        \sum_{
            \substack{
                p_{w}\in [0,W-1] \\ 
                p_{h}\in [0,H-1]}} (\hat{M_{ik}^{p_{w},p_{h}}}  -M_{ik}^{p_{w},p_{h}}))
    \end{equation}
\end{description}
To summarize, we assign higher importance to the data samples of the wrongly predicted instances in the incremental dataset.
This is done by increasing the weight so that it is in accordance with the obtained \ac{shap} value distances.
 
Thus, the weighting parameter $w_{i}$ of the sample $x_{i}$ is:
\begin{equation}
        w_{i} = \left\{ \begin{array}{rcl}
            1 & \mbox{if} & x_{i}\in D^{m} \\
            1+\Delta w_{i} & \mbox{if} & x_{i}\in D^{I_{1}} \cup \ldots D^{I_{s}} 
        \end{array} \right.
    \end{equation}

\section{Experiments}
\label{exp}

In this section, we first show the results of our method on a radar-based, real-life dataset. Afterward, to show the general purpose of our approach, we show how our approach work on the public computer vision dataset CIFAR-10.
\subsection{Experiments on Radar-based People Counting Dataset}
\subsubsection{Implementation Settings}
In the implementation step, we utilize mmWave \acrshort{fmcw} radar chipsets with one transmitter and three receiver units. 
			
		
		
		

			
We record zero to eight people moving around or standing still in an office environment using two radars. The recorded data was split into training, validation, and test sets. 
The training set consists of around 1.5M frames and 450K per validation and test set. We train the baseline network using the cross-entropy loss and the \ac{lar} loss without any data augmentation methods, with $N_{B}=18$ to align with the requirement from \cite{lar}. After training the baseline network, we start retraining the model by randomly shifting each frame from 0 to 5 ranges. Meanwhile, we select uniformly at randomly a subset of them (i.e., $10\%$) and flip them along the doppler axis.
The retraining sessions are repeated for ten consecutive incremental datasets. In each of these sessions, the model is trained in a stabilized manner and with a regular training strategy, with different weighting scenarios mentioned in Sec. \ref{weight}. After each session, we record more data to evaluate the network's performance and take the wrongly predicted data to create a new incremental dataset. Each incremental dataset contains around 15k frames. In the meanwhile, \ac{shap} values of these data are calculated. 

\subsubsection{Model Architecture}
\label{model architecture}
\begin{figure*}
    	\begin{center}
    	    \vspace{-0mm}
    	\includegraphics[width=0.85\textwidth]{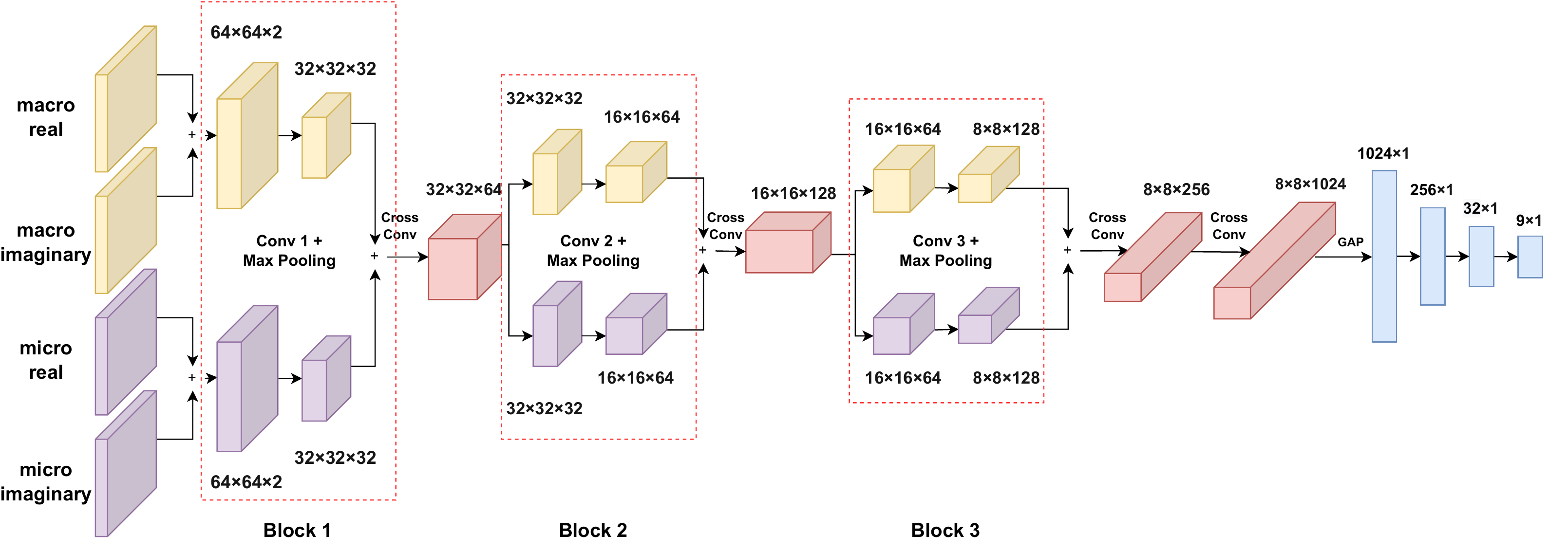}
    		\vspace{-2mm}
    	\end{center}
    	\caption{Network Architecture.}
    	\vspace{-4mm}
    	\label{fig:model}
\end{figure*}

Since the \ac{rdi} contains complex values, we can extract a real and imaginary part from it. 
In order to process  Macro- and Micro-information separately, we treat them as four different input channels: Macro-real map, Macro-imaginary map, Micro-real map, and Micro-imaginary map. As shown in Fig. \ref{fig:model}, we have two different pipelines for processing Macro- and Micro-chain. Each part consists of three blocks. The kernel size of convolution layers in each block is $5\times5$ with stride 1. After that, we also use \ac{bn} to increase the training speed and prevent overfitting. Immediately following a \ac{mp} layer with stride two can decrease the activation map size. After each block, we also concatenate Macro- and Micro-chains and further use Cross-Convolution to learn the connection between Macro- and Micro-information. 

In the Cross-Convolution stages, the kernel sizes are $2\times2$ to reduce computational costs. After the last Cross-Convolution stage, the size of the activation map is $8\times8\times1024$. We further use a \ac{gap} layer \cite{gap} to reduce the 3-dimensional map to a vector. At last, after Dense layers, we get 32-dimensional embedding vectors and 9-dimensional outputs for further prediction. The embedding vectors are used for \ac{dml} \cite{dml}, which aims to reduce the distance between samples from the same class and increase the distance between samples from different classes.

\subsubsection{Results}
 First, we compare our stability training architecture with \cite{stability}. Afterward, we show the result of retraining the network by giving weighting parameters to the data.

\begin{figure}[H]
\centering
     \includegraphics[width = 0.65\linewidth]{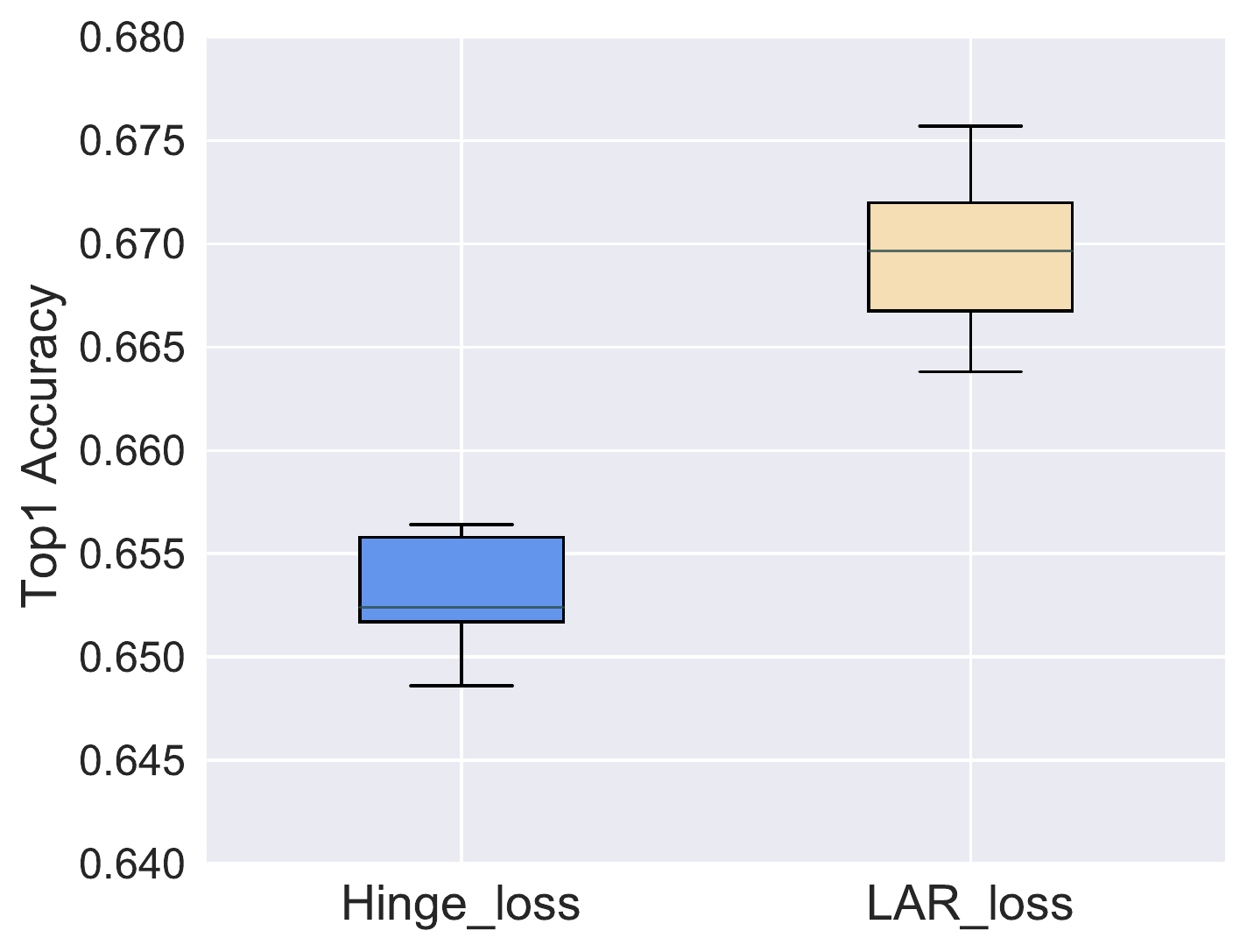}
     \caption{Stability Training Comparison.}
     \label{Fig: stability_compare}
     \vspace{-4mm}
 \end{figure}

\paragraph{Stability Training}
In order to minimize the influence brought by parameter initialization of the network, we train the two different stabilized architectures ten times each, with the same network settings, by augmenting the data from the training set $D^{m}$. Afterward, we evaluate the networks using the test set. The result is shown in Fig. \ref{Fig: stability_compare}. When we use \ac{lar} loss to stabilize the embedding vectors, the Top-1 accuracy reaches $66.96\%\pm 0.56\%$, which is $2\%$ more than hinge loss for deep metric learning. 

\begin{figure}
    \centering
    \begin{subfigure}[b]{.95\columnwidth}
        \includegraphics[width=\linewidth]{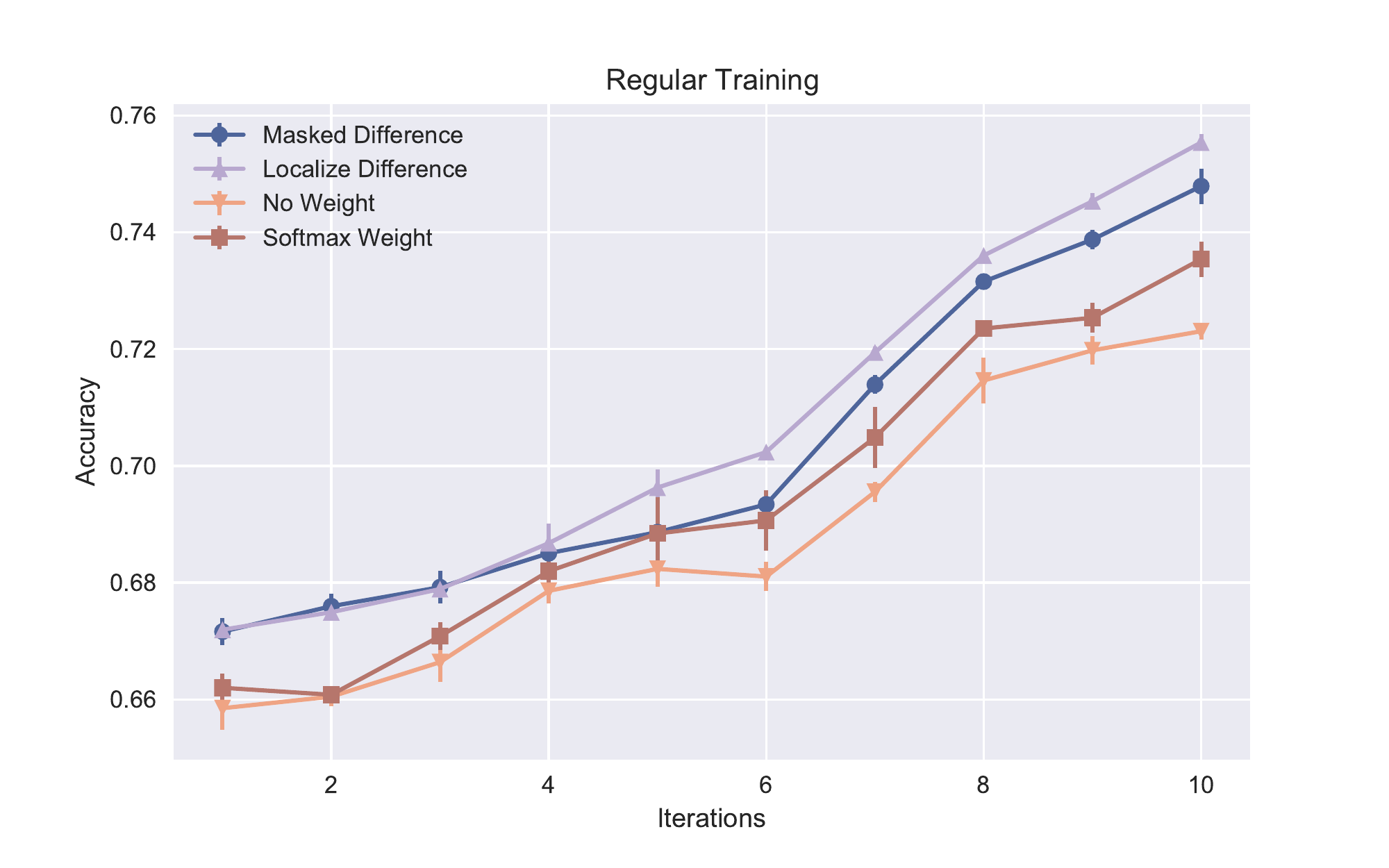}
        \caption{Regular Training.}
        \vspace{-0.2mm}
        \label{fig:regular_train}
    \end{subfigure}
    \hfill
    \begin{subfigure}[b]{.95\columnwidth}
        \includegraphics[width=\linewidth]{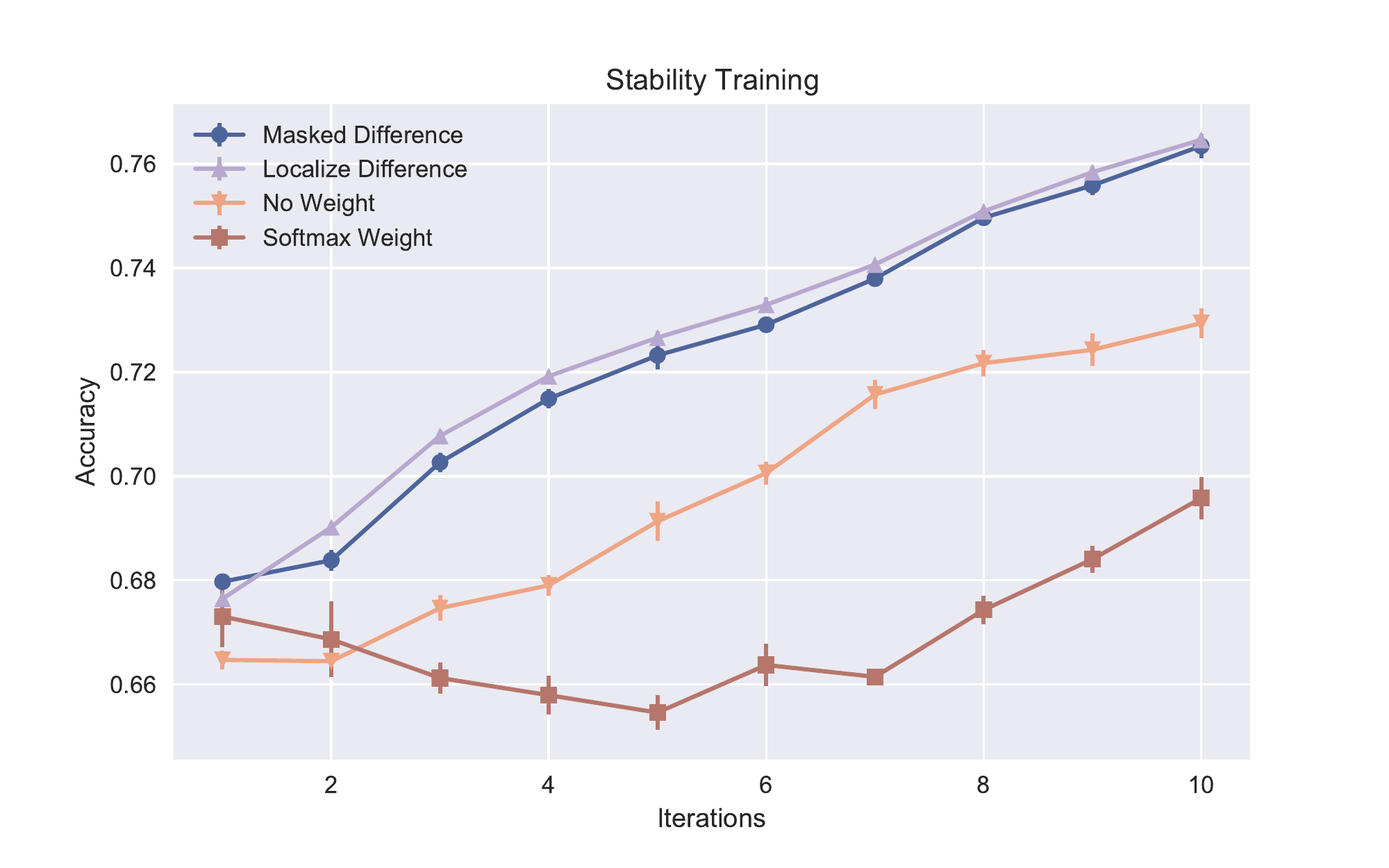}
        \caption{Stability Training.}
        \label{fig:stability_train}
        \vspace{-0.2mm}
    \end{subfigure}
\caption{Experiments Comparison.}
\setlength{\abovecaptionskip}{-0.2cm}   
\label{fig:exp}
\vspace{-4mm}
\end{figure}

\paragraph{Retraining Procedure}
We first train the baseline network with the original training set, and the Top-1 accuracy rate of the test set is $65.81\%$. We then split our retraining experiments into two parts: regular and stability training. We first retrain the network only with frequency-shifted methods in the regular training session. Then, we add Gaussian noise with a variance of 0.01 to the data and retrain the network with a stabilized architecture. The Gaussian variance is a hyper-parameter retrieved by using grid-search methods on the validation set.
This defines the stability of the training procedure. Afterward, we combine our weighting approaches with each one of the retraining experiments. This shows for the first time that \ac{xai} methods substantially improve the network performance by aligning the weighting parameter with the data sample. 
In order to establish a benchmark, we compare our weighting approaches to the non-weighted parameter retraining. This benchmark leads to eight different experiment combinations. In each experiment, we retrain the network incrementally ten sessions. In each of those sessions, we repeat the training for five runs to minimize the initialization randomness. The results are shown in Fig. \ref{fig:exp}.

Compared with no weight (i.e. equal weights of 1) and prediction-based weight retraining methods, our SHAP-based weighting methods improve the accuracy in both regular and stability training sessions. Overall, in the last retraining session, SHAP-based methods have a Top-1 accuracy rate of $3\%$ more than non-weighting retraining. Additionally, stability retraining with SHAP-based weighting parameters also helps to improve the network performance. However, calculating weights from the probability vector of predictions leads to an accuracy drop in the stability training scenario.

\subsection{Experiments on CIFAR-10}
In order to determine the generality of the proposed algorithm, we test it on CIFAR-10 \cite{cifar}, including 50k training samples and 10k test samples. We utilize 20k training data for the initial baseline training. The remaining 30k training data samples are further divided into four parts, creating the incremental datasets. To benchmark our approach, we use ResNet-18 \cite{resnet} structure. The baseline model has an $73.32\%$ accuracy rate on the test set. After retraining the baseline network four times by gradually adding new training samples, our methods end up with a more than $81.6\%$ accuracy rate, which is $3\%$ more than retraining with equal weight. The more detailed results are shown in Tab. \ref{table:regular_training}.

\begin{table*}[]
\centering
\setlength{\tabcolsep}{4pt}{
\begin{tabular}{c|cccc}
\hline
\multirow{1}{*}{}  
                  & Masked Difference & Localize difference & No Weight & Softmax Weight \\ \hline
Baseline          & \multicolumn{4}{c}{73.32\%}                                          \\ \hline
$1^{st}$ session          &  $76.69\% \pm 0.43\%$                &     $76.65\% \pm 0.32\%$                &   $76.02\% \pm 0.22\%$        &       $76.31\% \pm 0.36\%$                              \\ 
$2^{nd}$ session          &  $78.36\% \pm 0.39\%$                 &   $78.28\% \pm 0.42\%$                  &  $77.12\% \pm 0.45\%$          &     $77.09\% \pm 0.54\%$                                 \\ 
$3^{rd}$ session          &  $79.74\% \pm 0.36\%$                 &   $79.53\% \pm 0.43\%$                  &   $77.95\% \pm 0.56\%$         &    $77.68\% \pm 0.38\%$                                  \\ 
$4^{th}$ session          &    $\bm{81.94\% \pm 0.44\%}$               &   $\bm{81.86\% \pm 0.37\%}$                   &  $79.02\% \pm 0.36\%$          &   $78.99\% \pm 0.39\%$                                                              \\ \hline
\end{tabular}}
\caption{Results Comparison of retraining the baseline network four times with different methods on CIFAR-10.}
\label{table:regular_training}
\end{table*}
In addition, we also train the same structured network with all 50k SHAP-based weighted training samples. It achieves an $80.6\%$ accuracy rate on the test set, nearly $2\%$ less than retraining incrementally. Further shows that the model benefits from both SHAP-based weighting strategy and the proposed incremental retraining procedure.

\section{Conclusion}
\label{conclusion}

This paper proposes a retraining pipeline that assigns weights to misclassified samples by their \ac{shap} values. 
We conducted several experiments on data retrieved via short-range radar sensors as well as on a public dataset.
In order to establish the importance of the proposed approach, we first train a baseline network and successively start retraining sessions with different weighting methods. As a result, on the radar-based people counting dataset, we show that our SHAP-based weighting methods outperform retraining with standard equal weights of up to $4\%$. On the CIFAR-10, after four retraining sessions, the SHAP-based weighting method has an $3\%$ improved accuracy rate compared to the retraining procedure with equal weight and $2\%$ more than training from scratch using all the weighted data at once. 

\bibliographystyle{IEEEtran}
\bibliography{references}

\begin{thebibliography}{10}
\providecommand{\url}[1]{#1}
\csname url@samestyle\endcsname
\providecommand{\newblock}{\relax}
\providecommand{\bibinfo}[2]{#2}
\providecommand{\BIBentrySTDinterwordspacing}{\spaceskip=0pt\relax}
\providecommand{\BIBentryALTinterwordstretchfactor}{4}
\providecommand{\BIBentryALTinterwordspacing}{\spaceskip=\fontdimen2\font plus
\BIBentryALTinterwordstretchfactor\fontdimen3\font minus
  \fontdimen4\font\relax}
\providecommand{\BIBforeignlanguage}[2]{{%
\expandafter\ifx\csname l@#1\endcsname\relax
\typeout{** WARNING: IEEEtran.bst: No hyphenation pattern has been}%
\typeout{** loaded for the language `#1'. Using the pattern for}%
\typeout{** the default language instead.}%
\else
\language=\csname l@#1\endcsname
\fi
#2}}
\providecommand{\BIBdecl}{\relax}
\BIBdecl

\bibitem{you2019advanced}
C.~You, J.~Lu, D.~Filev, and P.~Tsiotras, ``Advanced planning for autonomous
  vehicles using reinforcement learning and deep inverse reinforcement
  learning,'' \emph{Robotics and Autonomous Systems}, vol. 114, pp. 1--18,
  2019.

\bibitem{grigorescu2020survey}
S.~Grigorescu, B.~Trasnea, T.~Cocias, and G.~Macesanu, ``A survey of deep
  learning techniques for autonomous driving,'' \emph{Journal of Field
  Robotics}, vol.~37, no.~3, pp. 362--386, 2020.

\bibitem{torres2018patient}
A.~D. Torres, H.~Yan, A.~H. Aboutalebi, A.~Das, L.~Duan, and P.~Rad, ``Patient
  facial emotion recognition and sentiment analysis using secure cloud with
  hardware acceleration,'' in \emph{Computational Intelligence for Multimedia
  Big Data on the Cloud with Engineering Applications}.\hskip 1em plus 0.5em
  minus 0.4em\relax Elsevier, 2018, pp. 61--89.

\bibitem{chen2020deep}
R.~Chen, L.~Yang, S.~Goodison, and Y.~Sun, ``Deep-learning approach to
  identifying cancer subtypes using high-dimensional genomic data,''
  \emph{Bioinformatics}, vol.~36, no.~5, pp. 1476--1483, 2020.

\bibitem{robotics1}
N.~S{\"u}nderhauf, O.~Brock, W.~Scheirer, R.~Hadsell, D.~Fox, J.~Leitner,
  B.~Upcroft, P.~Abbeel, W.~Burgard, M.~Milford \emph{et~al.}, ``The limits and
  potentials of deep learning for robotics,'' \emph{The International journal
  of robotics research}, vol.~37, no. 4-5, pp. 405--420, 2018.

\bibitem{robotics2}
H.~A. Pierson and M.~S. Gashler, ``Deep learning in robotics: a review of
  recent research,'' \emph{Advanced Robotics}, vol.~31, no.~16, pp. 821--835,
  2017.

\bibitem{samek2017explainable}
W.~Samek, T.~Wiegand, and K.-R. Müller, ``Explainable artificial intelligence:
  Understanding, visualizing and interpreting deep learning models,'' 2017.

\bibitem{cam}
B.~Zhou, A.~Khosla, A.~Lapedriza, A.~Oliva, and A.~Torralba, ``Learning deep
  features for discriminative localization,'' in \emph{Proceedings of the IEEE
  Conference on Computer Vision and Pattern Recognition (CVPR)}, June 2016.

\bibitem{gradcam}
\BIBentryALTinterwordspacing
R.~R. Selvaraju, M.~Cogswell, A.~Das, R.~Vedantam, D.~Parikh, and D.~Batra,
  ``Grad-cam: Visual explanations from deep networks via gradient-based
  localization,'' \emph{International Journal of Computer Vision}, vol. 128,
  no.~2, p. 336–359, Oct 2019. [Online]. Available:
  \url{http://dx.doi.org/10.1007/s11263-019-01228-7}
\BIBentrySTDinterwordspacing

\bibitem{lime}
M.~T. Ribeiro, S.~Singh, and C.~Guestrin, ``"why should i trust you?":
  Explaining the predictions of any classifier,'' 2016.

\bibitem{nlp1}
\BIBentryALTinterwordspacing
J.~Mullenbach, S.~Wiegreffe, J.~Duke, J.~Sun, and J.~Eisenstein, ``Explainable
  prediction of medical codes from clinical text,'' in \emph{Proceedings of the
  2018 Conference of the North {A}merican Chapter of the Association for
  Computational Linguistics: Human Language Technologies, Volume 1 (Long
  Papers)}.\hskip 1em plus 0.5em minus 0.4em\relax New Orleans, Louisiana:
  Association for Computational Linguistics, Jun. 2018, pp. 1101--1111.
  [Online]. Available: \url{https://aclanthology.org/N18-1100}
\BIBentrySTDinterwordspacing

\bibitem{nlp2}
\BIBentryALTinterwordspacing
A.~Amini, S.~Gabriel, S.~Lin, R.~Koncel-Kedziorski, Y.~Choi, and H.~Hajishirzi,
  ``{M}ath{QA}: Towards interpretable math word problem solving with
  operation-based formalisms,'' in \emph{Proceedings of the 2019 Conference of
  the North {A}merican Chapter of the Association for Computational
  Linguistics: Human Language Technologies, Volume 1 (Long and Short
  Papers)}.\hskip 1em plus 0.5em minus 0.4em\relax Minneapolis, Minnesota:
  Association for Computational Linguistics, Jun. 2019, pp. 2357--2367.
  [Online]. Available: \url{https://aclanthology.org/N19-1245}
\BIBentrySTDinterwordspacing

\bibitem{ppl_michael}
M.~Stephan, S.~Hazra, A.~Santra, R.~Weigel, and G.~Fischer, ``People counting
  solution using an fmcw radar with knowledge distillation from camera data,''
  in \emph{2021 IEEE Sensors}, 2021, pp. 1--4.

\bibitem{gesture}
S.~Hazra, H.~Feng, G.~N. Kiprit, M.~Stephan, L.~Servadei, R.~Wille, R.~Weigel,
  and A.~Santra, ``Cross-modal learning of graph representations using radar
  point cloud for long-range gesture recognition,'' \emph{arXiv preprint
  arXiv:2203.17066}, 2022.

\bibitem{tracking}
M.~Stephan, L.~Servadei, J.~Arjona-Medina, A.~Santra, R.~Wille, and G.~Fischer,
  ``Scene-adaptive radar tracking with deep reinforcement learning,''
  \emph{Machine Learning with Applications}, vol.~8, p. 100284, 2022.

\bibitem{radar_book}
A.~Santra and S.~Hazra, \emph{Deep learning applications of short-range
  radars}.\hskip 1em plus 0.5em minus 0.4em\relax Artech House, 2020.

\bibitem{xai_lr}
J.~h. Lee, I.~h. Shin, S.~g. Jeong, S.-I. Lee, M.~Z. Zaheer, and B.-S. Seo,
  ``Improvement in deep networks for optimization using explainable artificial
  intelligence,'' in \emph{2019 International Conference on Information and
  Communication Technology Convergence (ICTC)}, 2019, pp. 525--530.

\bibitem{xai_prun}
\BIBentryALTinterwordspacing
M.~Sabih, F.~Hannig, and J.~Teich, ``Utilizing explainable ai for quantization
  and pruning of deep neural networks,'' 2020. [Online]. Available:
  \url{https://arxiv.org/abs/2008.09072}
\BIBentrySTDinterwordspacing

\bibitem{cifar}
A.~Krizhevsky, G.~Hinton \emph{et~al.}, ``Learning multiple layers of features
  from tiny images,'' 2009.

\bibitem{ppl1}
C.~Y. Aydogdu, S.~Hazra, A.~Santra, and R.~Weigel, ``Multi-modal cross learning
  for improved people counting using short-range fmcw radar,'' in \emph{2020
  IEEE International Radar Conference (RADAR)}.\hskip 1em plus 0.5em minus
  0.4em\relax IEEE, 2020, pp. 250--255.

\bibitem{ppl2}
J.-H. Choi, J.-E. Kim, N.-H. Jeong, K.-T. Kim, and S.-H. Jin, ``Accurate people
  counting based on radar: Deep learning approach,'' in \emph{2020 IEEE Radar
  Conference (RadarConf20)}, 2020, pp. 1--5.

\bibitem{lar}
\BIBentryALTinterwordspacing
L.~Servadei, H.~Sun, J.~Ott, M.~Stephan, S.~Hazra, T.~Stadelmayer, D.~S.
  Lopera, R.~Wille, and A.~Santra, ``Label-aware ranked loss for robust people
  counting using automotive in-cabin radar,'' 2021. [Online]. Available:
  \url{https://arxiv.org/abs/2110.05876}
\BIBentrySTDinterwordspacing

\bibitem{shap}
\BIBentryALTinterwordspacing
S.~M. Lundberg and S.~Lee, ``A unified approach to interpreting model
  predictions,'' \emph{CoRR}, vol. abs/1705.07874, 2017. [Online]. Available:
  \url{http://arxiv.org/abs/1705.07874}
\BIBentrySTDinterwordspacing

\bibitem{shap_text}
\BIBentryALTinterwordspacing
W.~Zhao, T.~Joshi, V.~N. Nair, and A.~Sudjianto, ``{SHAP} values for explaining
  cnn-based text classification models,'' \emph{CoRR}, vol. abs/2008.11825,
  2020. [Online]. Available: \url{https://arxiv.org/abs/2008.11825}
\BIBentrySTDinterwordspacing

\bibitem{shap_time}
K.~E. Mokhtari, B.~P. Higdon, and A.~Ba{\c{s}}ar, ``Interpreting financial time
  series with shap values,'' in \emph{Proceedings of the 29th Annual
  International Conference on Computer Science and Software Engineering}, 2019,
  pp. 166--172.

\bibitem{gaussian}
S.~Dodge and L.~Karam, ``Understanding how image quality affects deep neural
  networks,'' in \emph{2016 eighth international conference on quality of
  multimedia experience (QoMEX)}.\hskip 1em plus 0.5em minus 0.4em\relax IEEE,
  2016, pp. 1--6.

\bibitem{adversarial}
C.~Szegedy, W.~Zaremba, I.~Sutskever, J.~Bruna, D.~Erhan, I.~Goodfellow, and
  R.~Fergus, ``Intriguing properties of neural networks,'' \emph{arXiv preprint
  arXiv:1312.6199}, 2013.

\bibitem{adversarial2}
S.-M. Moosavi-Dezfooli, A.~Fawzi, and P.~Frossard, ``Deepfool: a simple and
  accurate method to fool deep neural networks,'' in \emph{Proceedings of the
  IEEE conference on computer vision and pattern recognition}, 2016, pp.
  2574--2582.

\bibitem{radar_adversarial}
T.~Huang, Y.~Chen, B.~Yao, B.~Yang, X.~Wang, and Y.~Li, ``Adversarial attacks
  on deep-learning-based radar range profile target recognition,''
  \emph{Information Sciences}, vol. 531, pp. 159--176, 2020.

\bibitem{radar_adversarial2}
L.~Wang, X.~Wang, S.~Ma, and Y.~Zhang, ``Universal adversarial perturbation of
  sar images for deep learning based target classification,'' in \emph{2021
  IEEE 4th International Conference on Electronics Technology (ICET)}, 2021,
  pp. 1272--1276.

\bibitem{stability}
S.~Zheng, Y.~Song, T.~Leung, and I.~Goodfellow, ``Improving the robustness of
  deep neural networks via stability training,'' in \emph{Proceedings of the
  ieee conference on computer vision and pattern recognition}, 2016, pp.
  4480--4488.

\bibitem{triplet}
K.~Q. Weinberger, J.~Blitzer, and L.~Saul, ``Distance metric learning for large
  margin nearest neighbor classification,'' \emph{Advances in neural
  information processing systems}, vol.~18, 2005.

\bibitem{gap}
M.~Lin, Q.~Chen, and S.~Yan, ``Network in network,'' \emph{arXiv preprint
  arXiv:1312.4400}, 2013.

\bibitem{dml}
M.~Kaya and H.~{\c{S}}. Bilge, ``Deep metric learning: A survey,''
  \emph{Symmetry}, vol.~11, no.~9, p. 1066, 2019.

\bibitem{resnet}
\BIBentryALTinterwordspacing
K.~He, X.~Zhang, S.~Ren, and J.~Sun, ``Deep residual learning for image
  recognition,'' \emph{CoRR}, vol. abs/1512.03385, 2015. [Online]. Available:
  \url{http://arxiv.org/abs/1512.03385}
\BIBentrySTDinterwordspacing

\end{thebibliography}

\end{document}